\documentclass[10pt,conference]{IEEEtran}

\usepackage{cite}
\usepackage{amsmath,amssymb,amsfonts}
\usepackage{algorithmic}
\usepackage{graphicx}
\usepackage{textcomp}
\usepackage{xcolor}
\usepackage{tikz}   
\def\BibTeX{{\rm B\kern-.05em{\sc i\kern-.025em b}\kern-.08em
    T\kern-.1667em\lower.7ex\hbox{E}\kern-.125emX}}
\begin{document}

\title{Learning controllable dynamics through informative exploration}

\author{\IEEEauthorblockN{1\textsuperscript{st} Peter N. Loxley}
\IEEEauthorblockA{\textit{School of Science and Technology}}
\IEEEauthorblockA{\textit{University of New England}\\
Australia}
\and
\IEEEauthorblockN{2\textsuperscript{nd} Friedrich T. Sommer}
\IEEEauthorblockA{\textit{Redwood Center for Theoretical Neuroscience}}
\IEEEauthorblockA{\textit{University of California at Berkeley}\\
USA}
}

\maketitle

\begin{abstract}

Environments with controllable dynamics are usually understood in terms of explicit models. However, such models are not always available, but may sometimes be learned by exploring an environment. In this work, we investigate using an information measure called \emph{predicted information gain} to determine the most informative regions of an environment to explore next. Applying methods from reinforcement learning allows good suboptimal exploring policies to be found, and leads to reliable estimates of the underlying controllable dynamics. This approach is demonstrated by comparing with several myopic exploration approaches.

\end{abstract}


\section{Introduction}
Environments with controllable dynamics are interesting to understand and have numerous applications. In many cases we do not have available an explicit model of the environment as the controllable dynamics is generated by some black box; for example, a computer program that simulates the transition probabilities of an underlying Markov chain or the equations of motion of some noisy process. However, it is often desirable or even necessary to have an explicit model of the dynamics. One possibility is to learn the dynamics by exploring the environment. This is observed in animal behavior, where an inquisitive animal will often explore a novel environment by actively seeking information about the environment, enabling it to better prepare for future events such as avoiding predators \cite{woodgush,pisula}. More generally, active learning is the task of selecting what data to gather next so as to learn as much as possible \cite{mackay1992information}. Pfaffelhuber connected learning and information theory by proposing that learning is the process of decreasing missing information \cite{pfaffelhuber}. 

In this work, we apply informative exploration to learn the transition probabilities of a controllable Markov chain (CMC) in an unknown environment. More precisely, this work aims to determine a general method for finding a sequence of controls, called an exploring policy, to obtain the best estimate of an unknown CMC over a limited number of exploration steps. This is done within the framework of information theory and optimal control, building on the idea of optimal experimental design pioneered by Kristen Smith \cite{smith1918standard}, and further developed in works such as \cite{lindley1956measure, mackay1992information, oaksford1994rational,storck1995reinforcement,little,loxley}. The most closely related previous work appears in \cite{storck1995reinforcement} and \cite{little}. Here, we extend the work in \cite{storck1995reinforcement} by comparing with myopic exploration approaches, and by considering CMCs with transient and absorbing states -- making planning essential for good estimation. We extend both works by introducing a novel methodology for comparing estimated CMCs, and by applying finite-horizon methods that capture non-stationary exploring policies over short horizons. 

\section{Theory}
Environments with controllable dynamics are often described using controllable Markov chains (CMCs). Assuming a finite set of states $S$, and a finite set of controls $U(i)$ available at each state $i$, choosing control $u\in U(i)$ will move the environment from state $i$ to state $j$ according to the transition probabilities $p_{ij}(u)$. 

In many cases there is no explicit CMC available. A CMC can be estimated from data by counting the number of state transitions in the data, and storing them in a tensor $\boldsymbol{F}$; where tensor component $F_{uij}$ gives the number of transitions from state $i$ to state $j$ when control $u$ is chosen. A reasonable estimate of the underlying CMC is then given by $\widehat{p}_{ij}(u,\boldsymbol{F})=(F_{uij}+\alpha)/\sum_{j'}(F_{uij'}+\alpha)$ \cite{little}. This is the mean of a Dirichlet distribution with parameters $\boldsymbol{F}$ and $\alpha$ representing the posterior probability of each distribution $p_{i\cdot}(u)$, given the data $\boldsymbol{F}$ and a Dirichlet prior with parameter $\alpha$ \cite{mackay}.

To arrive at a good CMC estimate requires some form of informative exploration, visiting key states and transitions that have not previously been visited. Exploring an environment in an informative way can be done using an information measure called the \emph{predicted information gain} (PIG) \cite{little}. At each step this measure compares two alternative CMC estimates: the current CMC estimate, and an updated estimate found from exploring a new control. The greater the difference between these estimates, the larger the predicted information gain. The PIG is given by
\begin{equation}
\mathrm{PIG}(i,u,\boldsymbol{F})=\sum_{j^*} \widehat{p}_{ij^*}(u,\boldsymbol{F})\mathrm{D}_{\mathrm{KL}}[\widehat{p}_{i\cdot}(u,\boldsymbol{F}^{i\rightarrow j^*})||\widehat{p}_{i\cdot}(u,\boldsymbol{F}) ],
\end{equation}
where $\mathrm{D}_{\mathrm{KL}}$ is the KL divergence \cite{mackay} between the distributions $\widehat{p}_{i\cdot}(u,\boldsymbol{F})$ and $\widehat{p}_{i\cdot}(u,\boldsymbol{F}^{i\rightarrow j^*})$. The CMC estimate given by $\widehat{p}_{ij}(u,\boldsymbol{F})$ is simply the current estimate of $p_{ij}(u)$ based on the counts in $\boldsymbol{F}$. On the other hand, the CMC estimate given by $\widehat{p}_{ij}(u,\boldsymbol{F}^{i\rightarrow j^*})$ includes an additional exploring transition from $i$ to $j^*$ by temporarily updating $F_{uij}$ with an extra count to 
include this transition: denoted by $F_{uij}^{i\rightarrow j^*}$. It is important to note that PIG does not depend on $p_{ij}(u)$, which is unknown, but only on its estimate $\widehat{p}_{ij}(u,\boldsymbol{F})$.

The information measure given by PIG allows us to evaluate a control $u$ for its potential for exploring new states. We choose the control leading to the largest change in the current CMC estimate by maximizing PIG. This results from exploring new transitions not previously recorded in $\boldsymbol{F}$. The simplest approach to exploration with PIG is given by
\begin{equation}
\underset{u}{\operatorname{max}}\ \mathrm{PIG}(i,u,\boldsymbol{F}),\label{pigG}
\end{equation}
where we choose the control that maximizes the current PIG value without any regard to future PIG values during exploration -- this is called PIG greedy. An alternative greedy approach is given by
\begin{equation}
\underset{(i,u)}{\operatorname{max}}\ \mathrm{PIG}(i,u,\boldsymbol{F}),\label{upigG}
\end{equation}
where both state and control are chosen to jointly maximize the current PIG value -- this is called JPIG greedy (where the ``J" represents ``joint" optimization). 

Exploration involves both optimization and simulation. If the environment is in state $i$ during time period $k$, the maximization in (\ref{pigG}) leads to control $u$. Given the pair $(i,u)$, sampling $j\sim p_{i\cdot}(u)$ from the unknown transition probabilities is done via the black box (i.e., a computer program with input $(i,u)$ and output $j$), determining the next state $j$ at time period $k+1$. Then $F_{uij}$ is updated to $F_{uij}+1$, and the process repeats over a finite number of time periods. The choice of control $u$ during each time period determines the exploring policy -- represented abstractly as a sequence of functions $\pi=\{\mu_0,...,\mu_{N-1}\}$; where each function $\mu_k$ maps state $i$ at time period $k$ to control $u$: $\mu_k(i)=u$. Notice that updating $\boldsymbol{F}$ during each time period means $\mathrm{PIG}(i,u,\boldsymbol{F})$ is non-stationary, so that the exploring policy is also non-stationary (i.e., it depends on $k$). 

Making use of (\ref{upigG}) instead of (\ref{pigG}) modifies the exploration process. Now we ignore state $j$ after updating $F_{uij}$, instead finding the new state and control pair at time period \mbox{$k+1$} from the joint maximization in (\ref{upigG}). This has large consequences for exploration as the dynamics is no longer governed by $p_{ij}(u)$ via the black box, and so state $i_{k+1}$ is independent of state $i_k$. 

The exploration approaches using (\ref{pigG}) and (\ref{upigG}) are myopic: they focus on maximizing only the current value of PIG. This generally leads to suboptimal exploring policies. A better approach is to take future PIG values into account using \emph{dynamic programming} \cite{bert}. In this case, the finite-horizon dynamic programming algorithm takes the form:
\begin{equation}
J_k(i,\boldsymbol{F}) = \underset{u\in U_k(i)}{\operatorname{max}}\ \Big [\mathrm{PIG}(i,u,\boldsymbol{F})+\sum_j p_{ij}(u)J_{k+1}(j,F_{uij}+1)\Big ],\label{dp}
\end{equation}
where $F_{uij}$ is updated to $F_{uij}+1$ with probability $p_{ij}(u)$, corresponding to a transition from state $i$ to state $j$ under control $u$. This equation is iterated backwards from time period $k=N-1$ to time period $k=0$ for each possible state, with the maximizing $u$ yielding an optimal policy. Unfortunately, backward iteration of (\ref{dp}) suffers from Bellman's curse of dimensionality and is generally not possible. The reason is the `state' is now given by $(i,\boldsymbol{F})$, and since $\boldsymbol{F}$ depends on the entire history of exploration, the size of the corresponding state-space grows exponentially over time. 

A more natural approach for exploration is to approximate (\ref{dp}) using simulation. At time period $k$, the \emph{one-step lookahead approximation} of (\ref{dp}) at state $(i,\boldsymbol{F})$ is given by
\begin{equation}
\underset{u\in U_k(i)}{\operatorname{max}}\ \Big [\mathrm{PIG}(i,u,\boldsymbol{F})+\sum_j p_{ij}(u)\tilde{J}_{k+1}(j,F_{uij}+1)\Big ],\label{ro1}
\end{equation}
where $\tilde{J}_{k+1}$ is some approximation to $J_{k+1}$. If the environment is in state $i$ during time period $k$, the maximization in (\ref{ro1}) is used to find control $u$. To carry this out we use simulation. The sum in (\ref{ro1}) is approximated as a Monte Carlo average (via the black box simulator), and $\tilde{J}_{k+1}$ is approximated in a scheme known as \emph{rollout}\cite{bert}. This scheme requires a base policy, denoted by $\tilde{\pi}=\{\tilde{\mu}_0,...,\tilde{\mu}_{N-1}\}$, chosen to be the PIG greedy exploring policy in our case. The base policy allows us to simulate to the end of the problem, collecting the PIG at each time period along the way. The sum of these values is the value of $\tilde{J}_{k+1}$ in (\ref{ro1}). Details of this simulation are as follows. The equation for evaluation of the base policy is given by
\begin{align}
\tilde{J}_k(i,\boldsymbol{F}) &= \mathrm{PIG}(i,\tilde{\mu}_k(i),\boldsymbol{F})\nonumber\\
&+\sum_j p_{ij}\left (\tilde{\mu}_k(i) \right)\tilde{J}_{k+1}\left (j,F_{\tilde{\mu}_k(i)ij}+1 \right),\label{ro2}
\end{align}
and is approximated using simulation in the following way. During the first time period ($k=0$) all components of $\boldsymbol{F}$ are set to zero. During time period $k$ and at state $(i,\boldsymbol{F})$, add the value of $\mathrm{PIG}(i,\tilde{\mu}_{k}(i),\boldsymbol{F})$ to the sum of PIG values, draw a sample $j\sim p_{i\cdot}(\tilde{\mu}_k(i))$, and move to state $(j,F_{\tilde{\mu}_k(i)ij}+1)$. This is repeated until reaching the horizon at $k=N$. In the case of stochastic dynamics, the simulation is repeated a number of times to form a Monte Carlo average of $\tilde{J}_{k+1}$ in (\ref{ro1}), allowing the transition probabilities in (\ref{ro2}) to be properly represented. The policy generated using (\ref{ro1}) and (\ref{ro2}) is called a \emph{rollout policy}, and is guaranteed to be no worse than the base policy in the case of deterministic dynamics satisfying some mild technical conditions (sequential consistency, and ties broken in favour of the base policy) \cite{bert}. The rollout policy can also be used as the base policy for another rollout step, leading to a form of approximate policy iteration. This requires parameterizing policies in some way.

\section{Results}
We now estimate the simple CMC with two states and two controls shown in Fig.~1. For control $u=1$, the Markov chain has a transient state (state one) and an absorbing state (state two) provided $p<1$. For control $u=2$, each state is its own recurrent state. 
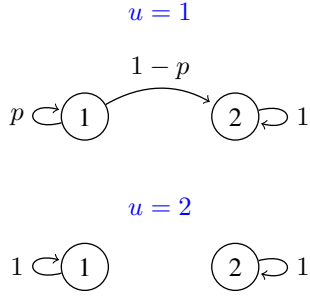
\begin{figure}
\centering
\begin{tikzpicture}[->,shorten >=2pt,line width=0.5pt,node distance=2cm]
\node [circle,draw] (one) {1};
\node [circle,draw] (two) [right of=one] {2};
\path (one) edge [loop left] node [left] {$p$} (one);
\path (one) edge [bend left] node [above] {$1-p$} (two);
\path (two) edge [loop right] node [right] {$1$} (two);
\node at (1,1.4) {\color{blue}{$u=1$}};
\node [circle,draw] (one-u2) [below of=one] {1};
\path (one-u2) edge [loop left] node [left] {$1$} (one-u2);
\node [circle,draw] (two-u2) [below of=two] {2};
\path (two-u2) edge [loop right] node [right] {$1$} (two-u2);
\node at (1,-1.2) {\color{blue}{$u=2$}};
\end{tikzpicture}
\caption{A controllable Markov chain (CMC) with two states and two controls.}
\end{figure}
The existence of a transient state makes this CMC difficult to estimate, as it requires choosing controls in a specific order to sample the transitions in state one. Doing so is a planning problem ideally suited to optimal control and sequential decision tasks. 

In Fig.~2, we estimate this CMC using several different exploration approaches by starting in state one and counting all transitions taking place while undergoing controllable dynamics. On the vertical axis, the missing information is the KL divergence between the two distributions $p_{i\cdot}(u)$ and $\widehat{p}_{i\cdot}(u,\boldsymbol{F})$, summed over all $i$ and $u$; with $\alpha=0.05$ to reflect our prior model ignorance. 
\begin{figure}
\centering
\includegraphics[scale=0.65,bb=100 300 450 550,clip=true]{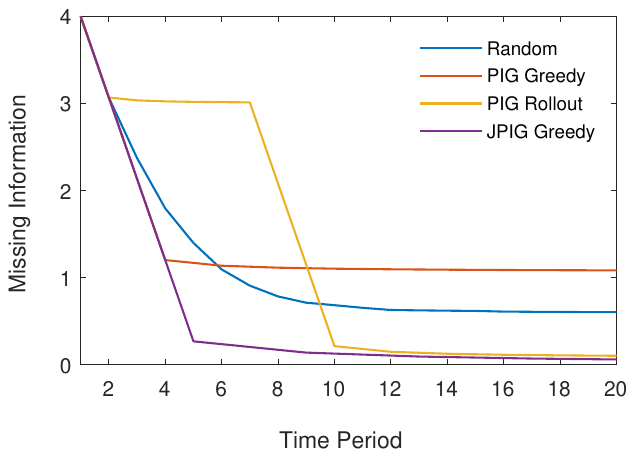}
\caption{Decrease in missing information over 20 time periods for four different approaches to estimating the CMC in Fig.~1 (with $p=0$) starting in state one.}
\label{}
\end{figure}
During random exploration, the control $u$ chosen uniformly at random from the control set $U_k(i)$ for an environment in state $i$ during time period $k$. Given the pair $(i,u)$ the simulation proceeds as previously described. In Fig.~2, random exploration is averaged over 200 trials, and is seen to be a reasonably effective exploration approach; doing better than PIG greedy, but not as well as JPIG greedy or PIG rollout. This result is somewhat surprising, as it contrasts with the environments investigated in \cite{little}, where random exploration was generally the worst performer. 

The exploring policies for PIG greedy and PIG rollout are shown in Fig.~3. For PIG greedy, its lack of planning ability means control one is selected and the environment immediately enters the absorbing state (state two) during the first time period. The result is that PIG greedy cannot access the transient state (state one) in order to sample its transitions, limiting the amount of missing information that can be learned. This policy then alternates between the two possible controls as the PIG objective ``encourages" new transitions to be explored. In Fig.~2, the steep decrease in missing information for PIG greedy during the first three time periods aligns with the jump between states, followed by the first two samples collected in state two (Fig.~3). We can now understand the surprising result for random exploration as due to the slightly longer time spent in the transient state, on average, than PIG greedy; allowing further sampling of the transitions from that state.

For PIG rollout, the exploring policy shown in Fig.~3 is due to its ability to ``look ahead" and carefully plan the choice of controls to optimize PIG. Choosing control two for the first six time periods allows sampling of the transient state, leading to better estimates of transitions from that state than for random exploration or PIG greedy. The PIG rollout policy then follows the PIG greedy policy for the remaining 2/3 of the time available, as seen from the steep decrease in missing information during time periods seven to nine in Fig.~2. Planning therefore gives PIG rollout a clear advantage over PIG greedy and random exploration in obtaining better samples of the CMC transition probabilities. 
\begin{figure}
\centering
\includegraphics[scale=0.75,bb=150 260 400 600,clip=true]{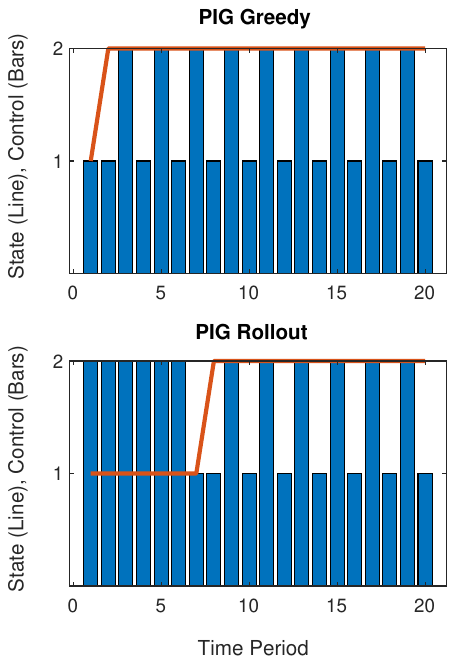}
\caption{Exploring policies for PIG greedy and PIG rollout corresponding to Fig.~2.}
\label{}
\end{figure}

Exploration for JPIG greedy is unconstrained by the state-dependency implicit in the $p_{ij}(u)$ transition probabilities, and freely alternates between each state and each control to explore new transitions according to the PIG objective. In Fig.~2, four time periods is long enough to allow both states and both controls to be visited once, leading to a steep decrease in missing information until reaching time period five. From Fig.~2, this leads JPIG greedy to the best CMC estimate, corresponding to the smallest missing information. As we shall now show, this exploration approach comes at a price. 

As the number of states of the environment increases, JPIG greedy will attempt to visit all states during exploration. However, in many real systems, controllable dynamics is concentrated on some unknown subset of states in a large state-space. The other exploration methods make use of the underlying CMC, which focuses sampling on the controllable dynamics. This is similar to the idea of using Markov chains in Monte Carlo methods to sample from high-dimensional probability distributions. A simple CMC that describes this situation is given in Fig.~4: states one and two form the same chain as in Fig.~1, while a number of additional states (labelled as $i$ and $j$) are included that have the same structure as states one and two for $u=2$. While somewhat contrived, this CMC captures the main idea. The controllable dynamics is similar to that shown in Fig.~1 when starting at state one, except that now the dynamics takes place on a small subset of states residing in a larger state-space. 
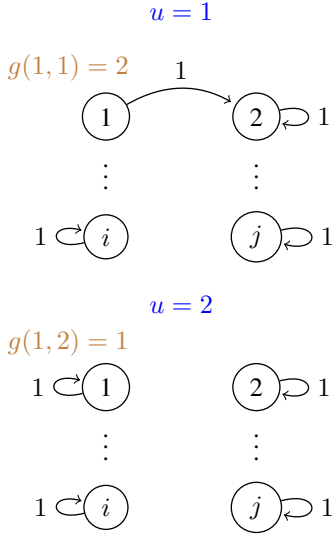
\begin{figure}
\centering
\begin{tikzpicture}[->,shorten >=2pt,line width=0.5pt,node distance=2cm]
\node [circle,draw] (one) {1};
\node [circle,draw] (two) [right of=one] {2};
\path (one) edge [bend left] node [above] {$1$} (two);
\path (two) edge [loop right] node [right] {$1$} (two);
\node at (1,1.4) {\color{blue}{$u=1$}};
\node at (-0.5,0.65) {\color{brown}{$g(1,1)=2$}};
\node at (0,-0.65) {.};
\node at (0,-0.8) {.};
\node at (0,-0.95) {.};
\node at (2,-0.65) {.};
\node at (2,-0.8) {.};
\node at (2,-0.95) {.};
\node [circle,draw] (one-u1) at (0,-1.6) {$i$}; 
\path (one-u1) edge [loop left] node [left] {$1$} (one-u1);
\node [circle,draw] (two-u1) at (2,-1.6) {$j$}; 
\path (two-u1) edge [loop right] node [right] {$1$} (two-u1);
\node at (1,-2.5) {\color{blue}{$u=2$}};
\node [circle,draw] (one-u2) [below of=one-u1] {1};
\path (one-u2) edge [loop left] node [left] {$1$} (one-u2);
\node [circle,draw] (two-u2) [below of=two-u1] {2};
\path (two-u2) edge [loop right] node [right] {$1$} (two-u2);
\node at (-0.5,-3) {\color{brown}{$g(1,2)=1$}};
\node at (0,-4.25) {.};
\node at (0,-4.4) {.};
\node at (0,-4.55) {.};
\node at (2,-4.25) {.};
\node at (2,-4.4) {.};
\node at (2,-4.55) {.};
\node [circle,draw] (i-u2) at (0,-5.2) {$i$}; 
\path (i-u2) edge [loop left] node [left] {$1$} (i-u2);
\node [circle,draw] (j-u2) at (2,-5.2) {$j$}; 
\path (j-u2) edge [loop right] node [right] {$1$} (j-u2);
\end{tikzpicture}
\caption{A CMC with many possible states over two controls, and controllable dynamics similar to that shown in Fig.~1. The controllable dynamics starting at state one takes place on a small subset of the full state-space.}
\end{figure}

We now estimate the CMC in Fig.~4 by exploring the environment. The decrease in missing information during exploration is shown in Fig.~5 for 100 states (states one and two, plus 98 additional states) over both controls. The plot looks very similar to Fig.~2, except the scale for missing information is different and the JPIG greedy curve has shifted up so that now PIG rollout gives the best exploring policy with the smallest missing information. Unlike in Fig.~2, JPIG greedy now visits the other states in the chain during time period five, and so the missing information in Fig.~5 stops decreasing (the missing information in Figs.~2 and 5 is calculated only for states one and two). In Ref.~\cite{little}, JPIG greedy (i.e., the unembodied agent) always gave the best exploring policy leading to the smallest missing information, while Fig.~5 shows that now PIG rollout does.
\begin{figure}
\centering
\includegraphics[scale=0.65,bb=100 300 450 550,clip=true]{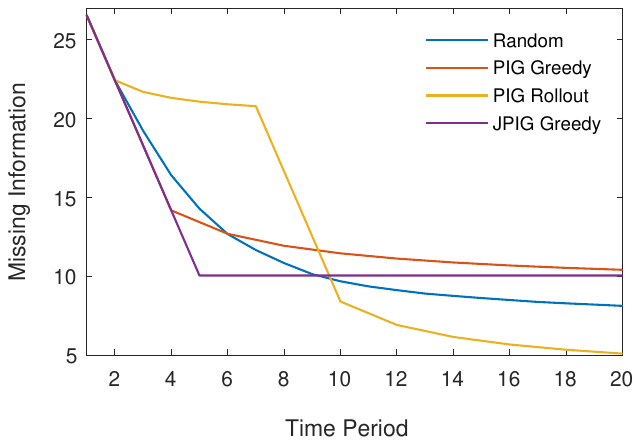}
\caption{Decrease in missing information over 20 time periods for four different approaches to estimating the CMC in Fig.~4 (with 100 states) starting in state one.}
\label{}
\end{figure}

Further detail is provided in Fig.~6 of the decrease in missing information as a result of sampling each transition from state one or state two. The most significant result is for $p_{11}(2)$ (top-left), where PIG rollout has the smallest missing information by a large margin. This takes place within the first six time periods, as confirmed in Fig.~3. All exploration approaches, except JPIG greedy, are comparable for the remaining transitions. As previously mentioned, JPIG greedy visits other states and transitions following time period four. Details of exploration at the level of individual transitions was not discussed in either \cite{storck1995reinforcement} or \cite{little}.
\begin{figure}
\centering
\includegraphics[scale=0.65,bb=100 280 500 550,clip=true]{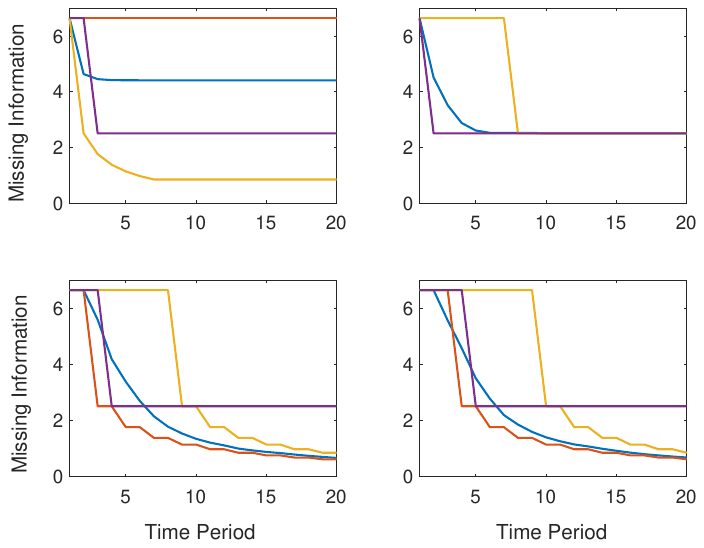}
\caption{Decrease in missing information as a result of sampling each transition from state one or state two in the CMC from Fig.~4. (Top-Left) $p_{11}(2)$, (Top-Right) $p_{12}(1)$, (Bottom-Left) $p_{22}(1)$, (Bottom-Right) $p_{22}(2)$. Curve labels are the same as in Figs.~2 and 5.}
\label{}
\end{figure}

We now compare the CMC estimates from each exploration approach by looking at their performance on a particular infinite horizon optimal control task. This task uses the same CMC as in Fig.~4, along with the following costs: $g(1,1)=2,g(1,2)=1$, and $g(i,u) = 0$ for all other $i$ and $u$. The optimal policy for this task is to exit the transient state as soon as possible (see Fig.~4). Under the true CMC, this can be done by choosing control one at a cost of two, then moving to state two for a cost of zero at all subsequent time periods. The total (discounted) cost of this policy is two. Under a CMC estimate, the optimal policy may choose control two for a cost of one, since the estimate $\widehat{p}_{12}(2,\boldsymbol{F})$ may be large enough to anticipate a move to state two for a cost of zero thereafter. However, this policy incurs a very large cost under the true CMC as the environment will remain in state one. On the other hand, if the estimate $\widehat{p}_{11}(2,\boldsymbol{F})$ is accurate enough, then $\widehat{p}_{12}(2,\boldsymbol{F})=1-\widehat{p}_{11}(2,\boldsymbol{F})$ will also be accurate enough, and this policy will be avoided. 

Introducing a discount factor $\alpha$, the corresponding Bellman equation\cite{bert} is given by
\begin{equation}
J^*(i) = \underset{u\in U(i)}{\operatorname{min}}\ \Big [g(i,u)+\alpha\sum_{j} \widehat{p}_{ij}(u,\boldsymbol{F})J^*(j)\Big ],\ \ \ \forall i,\label{bell}
\end{equation}
where $g(i,u)$ is the cost of state $i$ and control $u$. This equation can be solved using exact policy iteration. During each iteration, we evaluate the current policy by solving the following system of linear equations,
\begin{equation}
J_{\mu}(i) = g(i,\mu(i))+\alpha\sum_{j} \widehat{p}_{ij}(\mu(i),\boldsymbol{F})J_{\mu}(j),\ \ \ \forall i,\label{policyev}
\end{equation}
which in matrix-vector notation becomes,
\begin{equation}
MJ=g,\label{ls}
\end{equation}
where $M_{ij}=\delta_{ij}-\alpha\widehat{p}_{ij}(\mu(i),\boldsymbol{F})$, $J_i=J_{\mu}(i)$, and $g_i=g(i,\mu(i))$. After solving (\ref{ls}) for $J$, policy $\mu^k(i)$ can be ``improved" with the following step:
\begin{equation}
\mu^{k+1}(i) = \underset{\ u\in U(i)}{\operatorname{arg\ min}}\ \Big [g(i,u)+\alpha\sum_j \widehat{p}_{ij}(u,\boldsymbol{F})J_{\mu^k}(j)\Big ],
\end{equation}
where $J$ is used for $J_{\mu^k}(j)$. This process of policy evaluation, followed by policy improvement, is repeated until the policy converges: at which stage it is the optimal policy $\mu^*(i)$ solving the Bellman equation. Optimal policies solving the Bellman equation for each CMC estimate are given in Table 1.
\begin{table}[htbp]
\caption{Optimal policies for each CMC estimate are given in the middle column (where $i> 1$), with the discounted total cost ($\alpha=0.99$) of each policy under the true CMC given in the right column.}
\begin{center}
\begin{tabular}{l|l|l}
CMC Estimate&Optimal Policy&Total Cost under true CMC\\
\hline\hline
Random&$\mu^*(1)=2,\mu^*(i)=1$&$J(1)=100,J(i)=0$ \\
PIG Greedy&$\mu^*(1)=2,\mu^*(i)=1$&$J(1)=100,J(i)=0$ \\
JPIG Greedy &$\mu^*(1)=2,\mu^*(i)=1$&$J(1)=100,J(i)=0$\\
PIG Rollout &$\mu^*(1)=1,\mu^*(i)=1$&$J(1)=2,\ \ \ J(i)=0$
\end{tabular}
\end{center}
\end{table}

The discounted total cost of each policy \emph{under the true CMC} can be found using the method of policy evaluation:
\begin{equation}
J_{\mu^*}(i) = g(i,\mu^*(i))+\alpha\sum_{j} p_{ij}(\mu^*(i))J_{\mu^*}(j),\ \ \ \forall i.
\label{hyp}
\end{equation}
The discounted total cost $J_{\mu^*}(i)$ of each policy is also given in Table 1. From these results, it is clear the policy with the smallest total cost is given by PIG rollout, which is the same as the optimal policy for the true CMC. This means PIG rollout finds the best exploring policy for the CMC in Fig.~4.

Two exploring policies are shown for a more complex CMC with three states and three controls in Fig.~7. In this case, successful exploration requires choosing controls in a particular order to visit all three states. In Fig.~7, this is achieved by PIG rollout, but not by PIG greedy -- which only visits states one and three. 
\begin{figure}
\centering
\includegraphics[scale=0.75,bb=150 260 400 600,clip=true]{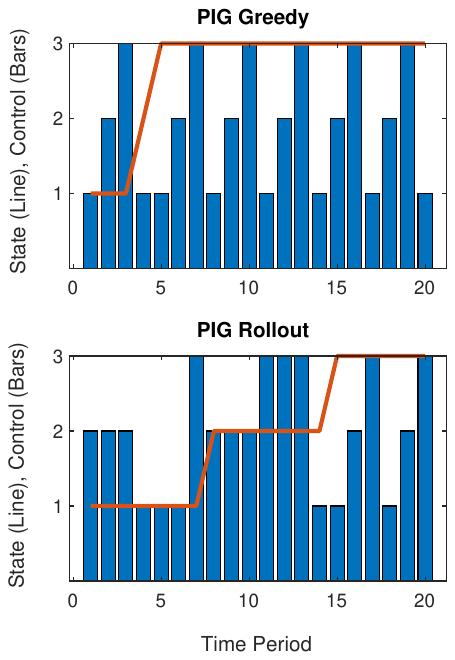}
\caption{Exploring policies for PIG greedy and PIG rollout for a CMC with three states and three controls. The rollout policy was improved using two policy iterations.}
\label{}
\end{figure}

\section{Conclusion}
In this work, information theory and optimal control methods were used to find policies for exploring an unknown environment and learning controllable dynamics. Algorithms from reinforcement learning and dynamic programming were shown to lead to better exploring policies than greedy algorithms, demonstrating a clear advantage of the ability to look ahead and plan, over myopic methods, when exploring an unknown environment. We extended previous work in \cite{storck1995reinforcement} and \cite{little} by providing a detailed analysis of the sampling dynamics and non-stationary policies for a nontrivial environment with transient and absorbing states. We also introduced a novel method for comparing model estimates of the controllable dynamics, motivated by the process of forming a hypothesis through exploration, then applying that knowledge to a specific task. In future work, we would like to further investigate this link between hypothesis formation and informative exploration. 

\bibliographystyle{IEEEtran.bst}
\bibliography{Loxley_references} 

\end{document}